\pdfoutput=1
\documentclass[11pt,a4paper,table]{article}
\usepackage[hyperref]{emnlp2020}

\usepackage{times}
\usepackage{latexsym}

\usepackage{microtype}

\aclfinalcopy % Uncomment this line for the final submission
\def\aclpaperid{74} %  Enter the acl Paper ID here

\usepackage{xcolor}
\usepackage[inline]{enumitem}

\usepackage[linesnumbered,ruled,vlined]{algorithm2e}
\usepackage{algorithmic}

\SetKwInput{KwInput}{Input}                % Set the Input
\SetKwInput{KwOutput}{Output}  
\usepackage{graphicx}
\usepackage[caption=false]{subfig}

\usepackage{booktabs}

\usepackage{array,multirow,graphicx}
 \usepackage{float}

\usepackage{amssymb} % for \smallsetminus
\usepackage{graphicx}
\usepackage{capt-of}% or \usepackage{caption}
\usepackage{booktabs}
\usepackage{varwidth}
\usepackage{multirow, booktabs}
\newsavebox\tmpbox

\newcommand{\eg}{e.g., }
\newcommand{\ie}{i.e., }

\newcommand{\figref}[1]{Fig.~\ref{#1}}    % within sentence
\newcommand{\tabref}[1]{Table~\ref{#1}}

\newcommand{\secref}[1]{Section~\ref{#1}}
\newcommand{\equref}[1]{Eq.~(\ref{#1})}
\newcommand{\task}{\mathcal{T}}
\newcommand{\loss}{\mathcal{L}}

\newcommand{\learner}{\texttt{M}}
\newcommand{\lossi}{\loss_{\task_i}}

\usepackage{blindtext}

\title{Zero-Shot Cross-Lingual Transfer with Meta Learning}

\author{Farhad Nooralahzadeh \\ University of Oslo \\ \texttt{farhad.nooralahzadeh@uzh.ch}
    \And Giannis Bekoulis \\ Vrije Universiteit Brussel - imec \\ \texttt{gbekouli@etrovub.be}
    \\\AND  Johannes Bjerva \\ University of Copenhagen / Aalborg University \\ \texttt{jbjerva@cs.aau.dk}
    \And Isabelle Augenstein \\University of Copenhagen  \\\texttt{augenstein@di.ku.dk}}

\date{}

\begin{document}
\maketitle
\begin{abstract}
Learning what to share between tasks has become a topic of great importance, as strategic sharing of knowledge has been shown to improve downstream task performance.
This is particularly important for multilingual applications, as most languages in the world are under-resourced.
Here, we consider the setting of training models on multiple different languages at the same time, when little or no data is available for languages other than English.
%when English training data, but little or no in-language data is available. %for languages other than English.
We show that this challenging setup can be approached using meta-learning: in addition to training a source language model, another model learns to select which training instances are the most beneficial to the first.
We experiment using standard supervised, zero-shot cross-lingual, as well as few-shot cross-lingual settings for different natural language understanding tasks (natural language inference, question answering).
Our extensive experimental setup demonstrates the consistent effectiveness of meta-learning for a total of 15 languages.
We improve upon the state-of-the-art for zero-shot and few-shot NLI (on MultiNLI and XNLI) and QA (on the MLQA dataset). %\todo{I added the MultiNLI dataset here. Please validate that the sentence is correct!}
A comprehensive error analysis indicates that the correlation of typological features between languages can partly explain when parameter sharing learned via meta-learning is beneficial.
\end{abstract}

\section{Introduction}
There are more than 7,000 languages spoken in the world, over 90 of which have more than 10 million native speakers each~\cite{ethnologue:19}.
Despite this, very few languages have proper linguistic resources when it comes to natural language understanding tasks \citep{joshi-etal-2020-state}.
Although there is growing awareness in the field, as evidenced by the release of datasets such as XNLI \cite{conneau2018xnli}, most NLP research still only considers English \cite{bender2019benderrule}.
While one solution to this issue is to collect annotated data for all languages, this process is both too time-consuming and expensive to be feasible.
Additionally, it is not trivial to train a model for a task in a particular language (\eg English) and apply it directly to another language where only limited training data is available (\ie low-resource languages).
Therefore, it is essential to investigate strategies that allow one to use the large amount of training data available for English for the benefit of other languages.

Meta-learning has recently been shown to be beneficial for several machine learning tasks \cite{koch:15,vinyals:16,santoro:16,finn:17,ravi:17,nichol:18}.
For NLP, recent work has also shown the benefits of this sharing between tasks and domains \cite{gu-etal-2018-meta,dou-etal-2019-investigating,qian-yu-2019-domain}.
Although cross-lingual transfer with meta-learning has been investigated for machine translation \cite{gu-etal-2018-meta}, this paper -- to best of our knowledge -- is the first attempt to study meta-learning for cross-lingual natural language understanding.
Our contributions are as follows:
\begin{itemize}[leftmargin=*]%[label=(\roman*)topsep=0pt]
    \item We propose X-MAML\footnote{Our code is available at \url{https://github.com/copenlu/X-MAML}}, a cross-lingual meta-learning architecture, and study it for two natural language understanding tasks (Natural Language Inference and Question Answering);
    \item We test X-MAML on cross-domain, cross-lingual, standard supervised, few-shot as well as zero-shot learning, across a total of 15 languages;
    \item We observe consistent improvements over strong models including Multilingual BERT~\citep{DBLP:journals/corr/abs-1810-04805} and XLM-RoBERTa~\citep{conneau2019unsupervised};
    %\item We observe consistent improvements with X-MAML over the state-of-the-art on various cross-lingual benchmarks (\ie improving our Multilingual BERT baseline model  by $+3.65\%$ and $+1.04\%$ points in terms of average accuracy on zero-shot and few-shot XNLI, respectively, and boosting our XLM-R$_{large}$ baseline  by $+1.47\%$ in terms of average F$_1$ score on zero-shot QA); and
     %(\ie an improvement over the state-of-the-art on zero-shot cross-lingual , as well as zero-shot and few-shot QA),  and
    \item We perform an extensive error analysis, which reveals that cross-lingual trends can partly be explained by typological commonalities between languages.
\end{itemize}

\section{Meta-Learning}
Meta-learning tries to tackle the problem of fast adaptation to a handful of new training data instances. It discovers the structure among multiple tasks such that learning new tasks can be done quickly. This is done by repeatedly simulating the learning process on low-resource tasks using many high-resource ones \cite{gu-etal-2018-meta}.
There are several ways of performing meta-learning: (i) metric-based~\cite{koch:15,vinyals:16}; (ii) model-based~\cite{santoro:16}; and (iii) optimisation-based~\cite{finn:17,ravi:17,nichol:18}. Metric-based methods aim to learn similarities between feature representations of instances from different training sets given a similarity metric. For model-based architectures, the focus has been on adapting models that learn fast (\eg memory networks) for meta-learning~\cite{santoro:16}. In this work, we focus on optimisation-based methods due to their superiority in several tasks (\eg computer vision~\cite{finn:17}) over the above-mentioned meta-learning architectures. These optimisation-based methods are able to find good initialisation parameter values and adapt to new tasks quickly.
To the best of our knowledge, we are the first to exploit the idea of meta-learning for transferring zero-shot knowledge in a cross-lingual setting for natural language understanding, in particular for the tasks of NLI and QA.
Specifically, we exploit the usage of Model Agnostic Meta-Learning (MAML) which uses gradient descent and achieves a good generalisation for a variety of tasks \cite{finn:17}. MAML is able to quickly adapt to new target tasks by using only a few instances at test time, assuming that these new target tasks are drawn from the same distribution. 

Formally, MAML assumes that there is a distribution $p(\mathcal{T})$ of tasks $\{\mathcal{T}_1, \mathcal{T}_2, ..., \mathcal{T}_k\}$. The parameters $\theta$ of model $\learner$ for a particular task $\mathcal{T}_i$, sampled from the distribution $p(\mathcal{T})$, are updated to ${\theta_i}^{'}$. In particular, the parameters $\theta$
are updated  using one or a few iterations of gradient descent steps on the training examples (\ie $D_{i}^{train}$) of task $\mathcal{T}_i$. For example, for one gradient update,
\begin{equation}\label{eq:fast-up}
    {\theta_i}^{'} = \theta - \alpha \nabla_\theta \mathcal{L}_{\mathcal{T}_i} (\learner_\theta)
\end{equation}
where $\alpha$ is the step size, the $\learner_\theta$ is the learned model from the neural network and $\mathcal{L}_{\mathcal{T}_i}$ is the loss on the specific task $\mathcal{T}_i$. The parameters of the model $\theta$ are trained to optimise the performance of $\learner_{\theta_i'}$ on the unseen test examples (\ie $D_{i}^{test}$) across tasks $p(\mathcal{T})$. The meta-learning objective is:
\begin{equation}
\min_\theta \sum_{\task_i \sim p(\task)}  \lossi ( \learner_{\theta_i'}) 
= \sum_{\task_i \sim p(\task)}  \lossi ( \learner_{\theta - \alpha \nabla_\theta \lossi(\learner_\theta)})
\end{equation}

\begin{algorithm*}[t]
%\small
\DontPrintSemicolon
 \KwInput{high-resource language \texttt{h}, set of low-resource languages \texttt{L},\\ Model $
 \learner$, step size $\alpha$ and learning rate $\beta$}
  
 Pre-train $\learner$ on \texttt{h} and provide initial model parameters $\theta$\\
 Select one or more languages from \texttt{L} as a set of auxiliary languages (\texttt{A}) \\
  \While{not done}
    { 
    \For{\texttt{l} $\in$ \texttt{A}} 
    {
    Sample batch of tasks $\task_i$ using the development set of the auxiliary language $l$ \\
    \For {each $\task_i$} 
    {
    Sample $K$ data-points to form $D_{i}^{train}=\{(X^{k},Y^{k})\}_{k=1}^{K} \in \task_i $\\
    Sample $Q$ data-points to form $D_{i}^{test}=\{(X^{q},Y^{q})\}_{q=1}^{Q} \in \task_i$~for meta-update\\
    Compute $\nabla_{\theta}\lossi(\learner_{\theta})$ on $D_{i}^{train}$ \\
    Compute adapted parameters with gradient descent:
    $\theta^{'}= \theta - \alpha\nabla_{\theta}\lossi(\learner_{\theta})$ \\
    Compute $\lossi(\learner_{\theta^{'}})$ using $D_{i}^{test}$  \\
    }
    }
    Update $ \theta \leftarrow \theta - \beta \nabla_{\theta} \sum_{i} \lossi(\learner_{\theta^{'}})$ \\
  }
Perform either (i) zero-shot or (ii) few-shot  learning on \{\texttt{L} $\smallsetminus$ \texttt{A}\} using meta-learned parameters $\theta$ 

\caption{X-MAML}
\label{alg:X-MAML}
\end{algorithm*}
The MAML algorithm aims to optimise the model parameters via a few number of gradient steps on a new task, which we
refer to as the meta-update.
The meta-update across all involved tasks is performed for the $\theta$ parameters of the model using stochastic gradient descent (SGD) as:
\begin{equation}
\label{eq:metaupdate}
\theta \leftarrow \theta - \beta \nabla_\theta \sum_{\task_i \sim p(\task)}  \lossi ( \learner_{\theta_i'})
\end{equation}
where $\beta$ is the meta-update step size.
\section{Cross-Lingual Meta-Learning}
\label{sec:mode_xmaml}
The underlying idea of using MAML in NLP tasks \cite{gu-etal-2018-meta,dou-etal-2019-investigating,qian-yu-2019-domain} is to employ a set of high-resource auxiliary tasks/languages to find an optimal initialisation from which learning a target task/language can be done using only a small number of training instances. In a cross-lingual setting (\ie XNLI, MLQA), where only an English dataset is available as a high-resource language, and a small number of instances are available for other languages, the training procedure for MAML requires some non-trivial changes.
For this purpose, we introduce a cross-lingual meta-learning framework (X-MAML), which uses the following training steps:
%[noitemsep]
\begin{enumerate}[leftmargin=*]
\setlength{\itemsep}{0pt}
\item Pre-training on a high-resource language \texttt{h} (\ie English): Given all training samples in a high-resource language \texttt{h}, we first train the model \texttt{M} on \texttt{h} to initialise the model parameters $\theta$. \label{step1}
\item Meta-learning using low-resource languages: This step consists of choosing one or more auxiliary languages from the low-resource set. Using the development set of each auxiliary language, we construct a randomly sampled batch of tasks $\task_i$. Then, we update the model parameters using $K$ data points of $\task_i$ ($D_{i}^{train}$) by one gradient descent step (see~\equref{eq:fast-up}). After this step, we can calculate the loss value using $Q$ examples ($D_{i}^{test}$) in each task.  It should be noted that the 
$K$ data points used for training ($D_{i}^{train}$) are different from the $Q$ data points used for testing ($D_{i}^{test}$). We sum up the loss values from all tasks to minimise the meta-objective function and to perform a meta-update using \equref{eq:metaupdate}. This step is performed in multiple iterations. \label{step2}
\item Zero-shot or few-shot learning on the target languages: In the last step of X-MAML, we first initialise the model parameters with those learned during
meta-learning. We then continue by evaluating the model on the test set of the target languages (\ie zero-shot learning) or fine-tuning the
model parameters with standard supervised learning using the development set of the target languages and evaluate on the test set (\ie few-shot learning). \label{step3}  
\end{enumerate}
A more formal description of the proposed model X-MAML is given in Algorithm \ref{alg:X-MAML}.

\paragraph{Natural Language Inference (NLI):}
NLI is the task of predicting whether a \textit{hypothesis} sentence is true (entailment), false (contradiction), or undetermined (neutral) given a \textit{premise} sentence.
%NLI is the task of inferring the logical relationship, typically entailment or contradiction, between a given \textit{hypothesis} and  \textit{premise} sentences.
The Multi-Genre Natural Language Inference (MultiNLI) dataset has 433k sentence pairs annotated with textual entailment information \cite{williams-etal-2018-broad}. 
It covers a range of different genres of spoken and written text and thus supports cross-genre evaluation. 
The NLI premise sentences are provided in 10 different genres: \textit{facetoface}, \textit{telephone}, \textit{verbatim}, \textit{state}, \textit{government}, \textit{fiction}, \textit{letters}, \textit{nineeleven}, \textit{travel} and \textit{oup}.
All of the genres appear in the test and development sets, but only five are included in the training set. To verify our learning routine more generally, we define $\task_i$ as an NLI task in each genre. We exploit MAML, in its original setting, to investigate whether meta-learning encourages the model to learn a good initialisation for all target genres, which can then be fine-tuned with limited supervision for each genre's development instances (2000 examples) to achieve a good performance on its test set.

The Cross-Lingual Natural Language Inference (XNLI)  dataset \cite{conneau2018xnli} consists of 5000 test and 2500 development hypothesis-premise pairs with their textual entailment labels for English. Translations of these pairs are provided in 14 languages: French (fr), Spanish (es), German (de), Greek (el), Bulgarian (bg), Russian (ru), Turkish (tr), Arabic (ar), Vietnamese (vi), Thai (th), Chinese (zh), Hindi (hi), Swahili (sw) and Urdu (ur). XNLI provides a multilingual benchmark to evaluate how to perform inference in low-resource languages, %such as Swahili or Urdu
in which only training data for the high-resource language English is available from MultiNLI.
%Following our X-MAML framework, we study the impact of meta-learning with one low-resource language to provide auxiliary tasks in the performance of a cross-lingual NLI model on the other target languages provided in the XNLI dataset. 
%The performance in NLI benchmarks is evaluated by accuracy on the test set.
%Following our X-MAML framework, we 
This allows us to study the impact of meta-learning with one low-resource language to serve as an auxiliary language, and evaluate the resulting NLI model on the target languages provided in the XNLI test set. %We evaluate the performance of a cross-lingual NLI model on the target languages provided in the XNLI dataset. 
%An evaluation on NLI benchmarks is performed reporting accuracy on the respective test sets.

\paragraph{Question Answering (QA):} Given a context and a question, the task in QA is to identify the span in the context which answers the question.
\newcite{lewis2019mlqa} introduce a Multilingual Question Answering dataset (MLQA)  that contains QA instances in 7 languages: English (en), Arabic (ar), German (de), Spanish (es), Hindi (hi), Vietnamese (vi) and Simplified Chinese (zh). It includes over 12k QA instances in English and 5k for every other language, with each QA instance being available in 4 languages (on average). This dataset has been used in many recent studies on cross-lingual transfer learning (\eg \citet{hu2020xtreme,Liang2020XGLUEAN}).
%For the evaluation, we use the F$_1$ score following the multilingual evaluation script available with the MLQA data\footnote{\url{https://github.com/facebookresearch/MLQA}}.
In our experiments, we investigate meta-learning for QA with one or two auxiliary languages.
 
\section{Experiments}
\label{sec:experiments}
%Our goal is to analyze the effect of X-MAML for various cross-lingual models.
%Therefore, we conduct experiments on the MultiNLI, XNLI and MLQA datasets using different models for the various tasks. 
We want to investigate how meta-learning can be used for cross-lingual sharing. %\footnote{Implementation notes are listed in Section \ref{setup}}
We implement X-MAML using the \textit{higher} library\footnote{\url{https://github.com/facebookresearch/higher}}. 
We use the Adam optimiser \cite{Adam} with a batch size of 32 for both zero-shot and few-shot learning. We fix the step size $\alpha$ and learning rate $\beta$ to $1e-4$ and $1e-5$, respectively. We experimented using  $[10,20,30,50,100,200,300]$ meta-learning iterations in X-MAML. However, $100$ iterations led to the best results in our experiments.  The sample sizes $K$ and $Q$ in X-MAML are equal to 16 for each dataset. The results are reported for each experiment by averaging the performance over ten different runs. We experiment with different architectures in order to verify that our method generalises across them, further detailed below. We report results for \textit{few-shot}, \textit{zero-shot} cross-domain and cross-lingual learning.
%\noindent
\paragraph{NLI:} We experiment with two different settings. (i) For MultiNLI, a cross-genre dataset, we employ the Enhanced Sequential Inference Model (ESIM) \cite{DBLP:journals/corr/ChenZLWJ16}, which is commonly used for textual entailment problems. ESIM uses LSTMs with attention to create a rich representation, capturing the relationship between premise and hypothesis sentences. 
(ii) For XNLI, a cross-lingual dataset, we use the PyTorch version of BERT~%\footnote{\url{https://github.com/huggingface/transformers}} 
using Hugging Face's library \cite{DBLP:journals/corr/abs-1810-04805} as the underlying model \texttt{M}. However, since our proposed meta-learning method is model-agnostic, it can easily be extended to any other architecture.
Note that for Setting (i), we apply MAML, whereas for Setting (ii), we apply X-MAML on the original English BERT model (En-BERT) and on Multilingual BERT (Multi-BERT) models. 
As the first training step (\ie pre-training on a high-resource language, see Step~\ref{step1} in~\secref{sec:mode_xmaml} for more information) in X-MAML for XNLI, we fine-tune En-BERT and Multi-BERT on the MultiNLI dataset (English) to obtain the initial model parameters $\theta$ for each experiment. 
\begin{table*}[t]
  \begin{varwidth}[b]{0.38\linewidth}
    \centering
        \fontsize{10}{12}\selectfont
\begin{tabular}{ccccc}
\toprule
& \textbf{Baseline} &\multicolumn{2}{c}{\textbf{MAML}}\\
\cmidrule(l){3-4}
\Large{x}\% & &  $\task_{Train}$ &  $\task_{Dev}$   \\
\midrule
\texttt{1}& 38.60 &	49.78 &	{\bf 50.92}  \\
\texttt{2} &37.80 &	48.58 &	{\bf 50.66}  \\
 \texttt{3} &47.09 &	51.40 &	{\bf 52.85}   \\
 \texttt{5} &49.88 &	{\bf 52.22} &	51.40   \\
 \texttt{10} &51.02 &	52.51 &	{\bf 53.95}  \\
 \texttt{20} &59.14&	{\bf 61.38} &	58.16  \\
 \texttt{50}&63.37&	{\bf 63.85} &
 61.74  \\
 \texttt{100}&64.35&{\bf 64.99} &64.61  \\
\bottomrule
\end{tabular}
        \caption{Test accuracies with different settings of MAML on MultiNLI. \texttt{x}\%: the percentage of training samples. {\bf Baseline:} The test accuracy of trained ESIM using \texttt{x}\% of training data. {\bf MAML:} The test accuracy of ESIM after meta-learning, where $\task_{Train}$: 5 tasks are defined in MAML using the training set, and $\task_{Dev}$: 10 tasks are included in MAML using the development set. Bold font indicates best results for the various proportions of the used training data.}
\label{tbl:mnli}
  \end{varwidth}%
  \hfill
  \begin{minipage}[b]{0.59\linewidth}
    \centering
    \includegraphics[width=1.0\textwidth]{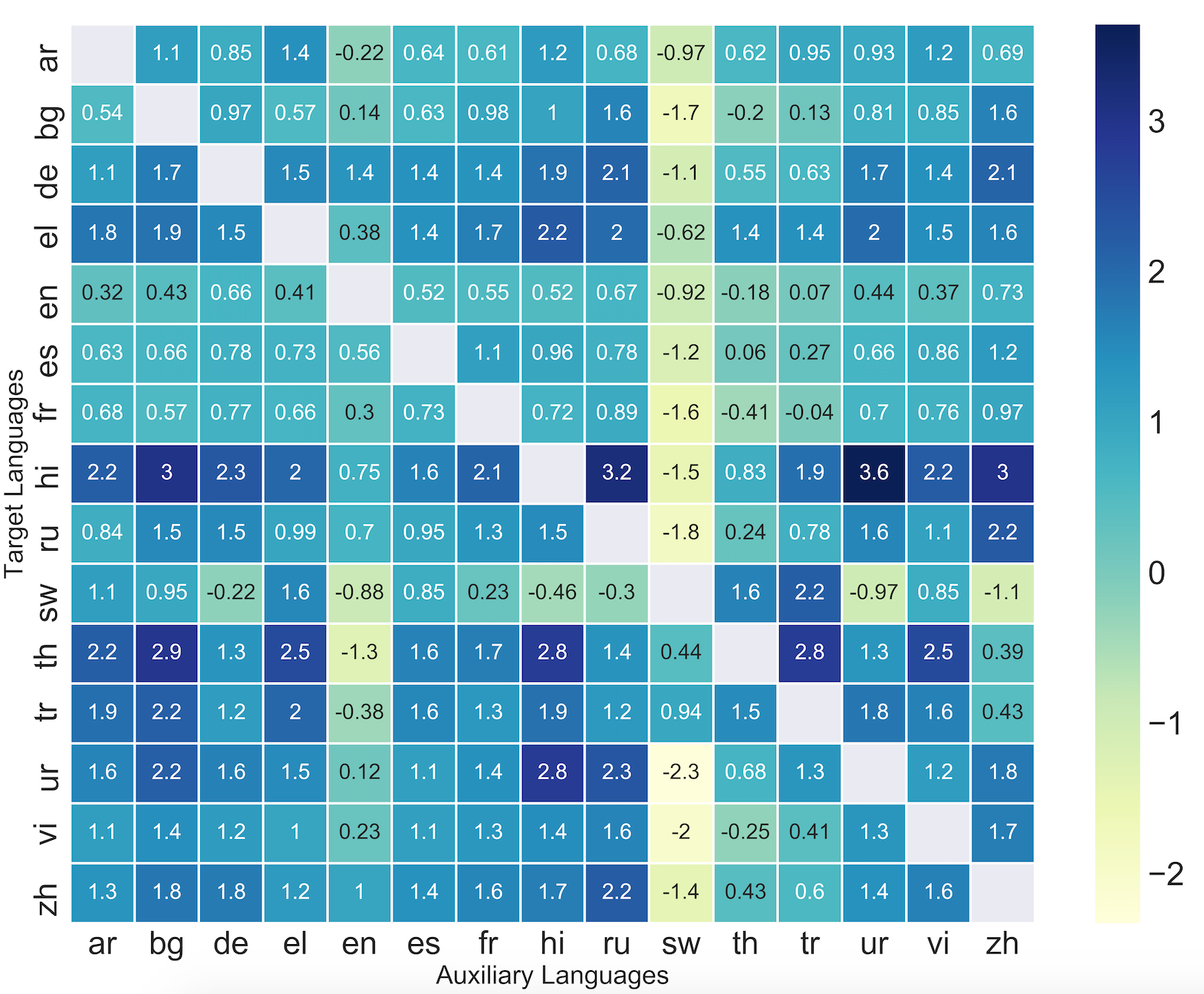}
    \captionof{figure}{Differences in performance in terms of accuracy scores on the test set for zero-shot X-MAML on XNLI using the Multi-BERT model. Rows correspond to target and columns to auxiliary languages used in X-MAML. Numbers on the off-diagonal indicate performance differences  between X-MAML and the baseline model in the same row. The coloring scheme indicates the differences in performance (\eg blue for large improvement).}
    \label{fig:heat-map-mBERT-zero}
  \end{minipage}
\end{table*}

\paragraph{QA:} 
%In order to validate whether our framework is model- and task- agnostic and can be applied to other tasks and models besides NLI and BERT, we conduct additional experiments on the QA task. 
For question answering, we use different base models \texttt{M} for X-MAML, namely XLM~\citep{conneau2019cross} and XLM-RoBERTa (XLM-R)~\citep{conneau2019unsupervised}, both state-of-the-art models.
%We use these models as the base model \texttt{M} in X-MAML for our QA experiments.
XLM uses a similar pre-training objective as Multi-BERT with a larger model, a larger
shared vocabulary, and leverages both monolingual and parallel data. XLM-R is a RoBERTa %(Liu et al., 2019)-
version of XLM and is trained on a much larger multilingual corpus (\ie Common Crawl), achieving state-of-the-art performance on most cross-lingual benchmarks \citep{hu2020xtreme}. 
We employ the XLM-15 (Masked Language Model + Translation Language Model, 15 languages), XLM-R$_{base}$ and XLM-R$_{large}$ models released by the authors.
The SQuAD v1.1 training data is used in the pre-training step of X-MAML (see Step~\ref{step1} in ~\secref{sec:mode_xmaml}). We use the cross-lingual development and test splits provided in the MLQA dataset for the meta-learning and evaluation steps, respectively.

\paragraph{Baselines:}
We create: 
\begin{enumerate*}[label=(\roman*)]
\item zero-shot baselines: directly evaluate the model on the test set of the target languages;
\item few-shot baselines: fine-tune the model on the development set, then evaluate on the test set of the low-resource languages.
\end{enumerate*}

\subsection{Few-Shot Cross-Domain NLI}
We train ESIM on the MultiNLI training set to provide initial model parameters $\theta$ (see Step~\ref{step1} in~\secref{sec:mode_xmaml}). We evaluate the pre-trained model on the English test set of XNLI (since the MultiNLI test set is not publicly available) as a baseline. 
Since MultiNLI is already split into genres, we use each genre as a task within MAML. 
We then include either the training set (5 genres) or the development set (10 genres) during meta-learning (similar to Step~\ref{step2} in X-MAML).  

In the last phase (similar to Step~\ref{step3} in X-MAML), we first initialise the model parameters with those learned  by MAML. We then continue to fine-tune the model using the development set of MultiNLI and report the accuracy on the English test set of XNLI.
We proportionally select sub-samples $x=[1\%,2\%,3\%,5\%,10\%,20\%,50\%,100\%]$ from the training data (with random sampling).

The results obtained by training on the corresponding proportions ($x\%$) of the MultiNLI dataset using ESIM (as the learner model \texttt{M}) are shown in \tabref{tbl:mnli}. We observe that for both settings (\ie MAML on the training (5 tasks) and on the development set (10 tasks)), the performance of all models (including baselines) improve as more instances become available. However, %as demonstrated by our experimental study, 
the effectiveness of MAML is larger when only limited training data is available (improving by $12\%$ in accuracy when $2\%$ of the data is available on the development set). 
\addtolength{\tabcolsep}{-4pt} % hack to have thinner columns
\begin{table*}[t]
\begin{center}
\resizebox{1\linewidth}{!}{
\begin{tabular}[b]{p{4.3cm}|ccccccccccccccc|c}
\toprule
& \textbf{en} & \textbf{fr} & \textbf{es} & \textbf{de} & \textbf{el} & \textbf{bg} & \textbf{ru} & \textbf{tr} & \textbf{ar} & \textbf{vi} & \textbf{th} & \textbf{zh} & \textbf{hi} & \textbf{sw} & \textbf{ur} & \textbf{avg} \\
\midrule
\rowcolor{gray!20}\multicolumn{16}{l}
{\it  Zero-shot cross-lingual transfer} \\ 
\midrule
\newcite{DBLP:journals/corr/abs-1810-04805} &81.4& - &74.3 &70.5 &- &-& -& - &62.1& -& -& 63.8 &- &-& 58.35 & - \\
\newcite{DBLP:journals/corr/abs-1904-09077} &  82.1 & 73.8 & 74.3 & 71.1 & 66.4 & 68.9 & 69.0 & 61.6 & 64.9 & 69.5 &  55.8 & 69.3 & 60.0 & 50.4 & 58.0  & 66.3 \\
Multi-BERT (Our baseline) & 81.36 & 73.45&73.85& 69.74& 65.73 & 67.82& 67.94 & 59.04 & 64.63 & 70.12 & 52.46 & 68.90 & 58.56 & 47.58 & 58.70 & 65.33 \\
\midrule
\multicolumn{16}{l}{\it \textbf{X-MAML} (One aux. lang.)} \\ 
\midrule
AVG & 81.69	& 73.86 &	74.43	& 71.00 &	67.16 &	68.39 &	68.90 &	60.41 &	65.33 &	70.95 &	54.08 &	70.09	& 60.51&	47.97&	59.94 & -\\
MAX & 82.09 &	 74.42 &	 75.07 &	 71.83 &	67.95 &	 69.45 &	 70.19 &	61.20 &	 66.05&	{ 71.82}&	55.39&	{ 71.11}&	{ 62.20} &	49.76 &	{61.51}&{ 67.33} \\
$hi\rightarrow X$ & 81.88	& 74.17 &	74.81 &	71.59 &	67.95 &	68.86 & 69.44 &	60.93 &	65.86 &	71.57 &	55.26 &	70.59 &	- &	47.12 &	61.51 & - \\
\midrule
{\it \textbf{X-MAML} (Two aux. lang.)} & {\it (hi,de)} & {\it (hi,ar)} & {\it (fr,de)} & {\it (bg,zh)} & {\it (ur,ru)} & {\it (hi,ru)} & {\it (de,bg)} & {\it (ur,sw)} & {\it (el,tr)} & {\it (de,bg)} & {\it (bg,tr)} & {\it (ru,el)} & {\it (ur,ru)} & {\it (el,tr)} & {\it (hi,de)}\\ 
\midrule
$(l_1,l_2) \rightarrow X$& {\bf 82.59} & {\bf 75.69} & {\bf 75.97} & {\bf 73.45} & {\bf 69.16} & {\bf 71.42} & {\bf 71.44}& {\bf 62.57} & {\bf 67.19} & {\bf 72.63} & {\bf 62.57} &{\bf 73.13} & {\bf 63.53} & {\bf 50.42} & {\bf 62.93} & {\bf 68.98}
\\
\midrule
\midrule
\rowcolor{gray!20}\multicolumn{16}{l}{\it  Few-Shot learning} \\
\midrule
Multi-BERT (Our baseline) & 81.94 & {  75.39} & 75.79 & 73.25 & 69.54 &71.60 &70.84 & 64.85 &67.37 &73.23 & 61.18  & 73.93  &  64.37 & 57.82  &  63.71   &  69.65 \\
\midrule
\multicolumn{16}{l}{\it \textbf{X-MAML} (One aux. lang.)} \\ 
\midrule
AVG & 82.22&	75.24&	76.06	&73.34&	69.97&	71.80&	71.28&	64.76&	67.82&	73.41&	61.57&	74.02&	64.83&	58.02&	63.66& -\\ 
MAX& { 82.39}	&  75.32& 	{  76.18}	& {  73.46}& 	{  70.03}& 	{  71.94}& 	{  71.45}& 	{  64.92}& 	{  67.95}& 	{  73.52}& 	{  61.74}& 	{  74.21}& {  	64.97}& 	{  58.23}& 	{  63.81} & { 70.01}\\
$sw\rightarrow X$ & 82.24	&75.31&	75.94&	73.34&	69.98&	71.77&	71.31&	64.89&	67.87	&73.38&	61.5&	73.99&	64.94&	- &	63.63 & -\\ % avg=70.72
\midrule
{\it \textbf{X-MAML} (Two aux. lang.)} & \textit{ (ar,ru)}&\textit{ (ru,th)}&\textit{ (ru,th)}&\textit{ (el,hi)}&\textit{ (sw,vi)}&\textit{ (ar,zh)}&\textit{ (de,tr)}&\textit{ (es,sw)}&\textit{ (bg,hi)}&\textit{ (bg,ru)}&\textit{ (el,vi)}&\textit{ (ar,th)}&\textit{ (sw,vi)}&\textit{ (ar,tr)}&\textit{ (en,ru)}\\ 
\midrule
$(l_1,l_2) \rightarrow X$& {\bf 82.71} & 75.97 & 76.51 & 74.07 & 70.66 & 72.77 & 72.12 & 65.69 & 68.4 & {\bf 73.87} & 62.5 & 74.85 & {\bf 65.75} & 59.94 & {\bf 64.59} & 70.69
\\
\midrule
\rowcolor{gray!20}\multicolumn{16}{l}{\it Machine translate at test (TRANSLATE-TEST)} \\
\midrule
\newcite{DBLP:journals/corr/abs-1810-04805} & 81.4& - &74.9 &74.4 &- &-& -& -& 70.4& -& -& 70.1& - &- &62.1 & -\\
\midrule
\rowcolor{gray!20}\multicolumn{16}{l}{\it Machine translate at training (TRANSLATE-TRAIN)} \\
\midrule
\newcite{DBLP:journals/corr/abs-1904-09077} &  82.1&	{\bf 76.9}&	{\bf 78.5}&	{\bf 74.8}&	{\bf 72.1}&	{\bf 75.4}&	{\bf 74.3}&	{\bf 70.6}&	{\bf 70.8}&	67.8&	{\bf 63.2}&	{\bf 76.2}&	65.3&	65.3&	60.6 & {\bf 71.6} \\
\bottomrule
\end{tabular}
}
\caption{Accuracy results on the XNLI test set for zero- and few-shot X-MAML. Columns indicate the target languages. %Multi-BERT \cite{DBLP:journals/corr/abs-1810-04805} is the original model introduced in \cite{DBLP:journals/corr/abs-1810-04805}; Multi-BERT (Wu) is reported by \newcite{DBLP:journals/corr/abs-1904-09077} where they reproduce Multi-BERT on all languages. 
The models of~\citet{DBLP:journals/corr/abs-1810-04805} and \citet{DBLP:journals/corr/abs-1904-09077} are also Multi-BERT models. For our Multi-BERT baseline model for (i) zero-shot learning, we evaluate the pre-trained model on the test set of the target language; and for (ii) few-shot learning, we fine-tune the model on the development set and evaluate on the test set of the target language. The avg column indicates row-wise average accuracy. We also report the average (AVG) and maximum (MAX) performance by using one auxiliary language for each target language.
$(l_1,l_2)$ are the most beneficial auxiliary languages for X-MAML in improving the test accuracy of each target language $X$. In TRANSLATE-TEST~\cite{DBLP:journals/corr/abs-1810-04805}, the target language test data is translated to English and then the model is fine-tuned on English. In TRANSLATE-TRAIN~\cite{DBLP:journals/corr/abs-1904-09077}, the English training data is translated to the target language and the model is fine-tuned using the translated data.} 
\label{tab:originalBERT}
\end{center}
\end{table*}

\subsection{Zero- and Few-Shot Cross-Lingual NLI}
\paragraph{Zero-Shot Learning:} 
In this set of experiments, we employ the proposed framework (\ie X-MAML) within a zero-shot setup, in which we do not fine-tune after the meta-learning step. 
We report the impact of meta-learning for each target language as a difference in accuracy with and without meta-learning on top of the baseline model (Multi-BERT) on the test set (\figref{fig:heat-map-mBERT-zero}). Each column corresponds to the performance of Multi-BERT after meta-learning with a single auxiliary language, and evaluation on the target language of the XNLI test set. Overall, we observe that our zero-shot approach with X-MAML outperforms the baseline model without MAML and results reported by \newcite{DBLP:journals/corr/abs-1810-04805}. This way, we improve the state-of-the-art performance for zero-shot cross-lingual NLI (in several languages for up to $+3.6\%$ in accuracy, \eg Hindi (hi) as target and Urdu (ur) as auxiliary language). For the exact accuracy scores, we refer to \tabref{tab:xnli-mBERT-zeroshot} in the Appendix. We hypothesise that the degree of typological commonalities among the languages has an effect (\ie positive or negative) on the performance of X-MAML. It can be observed that the proposed learning approach provides positive impacts across most of the target languages. However, including Swahili (sw) as an auxiliary language in X-MAML is not beneficial for the performance on the other target languages. It is worth noting that we experimented by just training the model using an auxiliary language, instead of performing meta-learning (step~\ref{step2}). From this experiment, we observe that meta-learning has a strongly positive effect on predictive performance (see also~\figref{fig:heatmap-NORMAL-MBERT} in the Appendix).

In Table \ref{tab:originalBERT}, we include the original baseline performances reported in \newcite{DBLP:journals/corr/abs-1810-04805}\footnote{\url{https://github.com/google-research/bert/blob/master/multilingual.md}} and \newcite{DBLP:journals/corr/abs-1904-09077}. We report the average and maximum performance by using one auxiliary language for each target language. We also report the performance of X-MAML by also using Hindi (which is the most effective auxiliary language for the zero-shot setting, as shown in ~\figref{fig:heat-map-mBERT-zero}). We suspect that this is because of the typological similarities between Hindi (hi) and other languages. 
Furthermore, by using two auxiliary languages in X-MAML results to the largest benefit in our zero-shot experiments.
\paragraph{Few-Shot Learning:}
For few-shot learning, meta-learning in X-MAML (Step~\ref{step3}) is performed by fine-tuning on the development set (2.5k instances) of target languages, and then evaluating on the test set. Detailed ablation results are presented in the Appendix (\tabref{tab:xnli-mBERT-fewshot} and \figref{fig:ffew-shot-heatmap-MAML-MBERT}). In~\tabref{tab:originalBERT}, we compare X-MAML results with one or two auxiliary languages to the external and internal baselines. We also showcase the performance using specifically Swahili (sw), the overall most effective auxiliary language for meta-learning with Multi-BERT in the few-shot learning setting. In addition, we report results from~\newcite{DBLP:journals/corr/abs-1810-04805} that use machine translation at test time (TRANSLATE-TEST) and results from~\newcite{DBLP:journals/corr/abs-1904-09077} that use machine translation at training time (TRANSLATE-TRAIN). Note that, using X-MAML, we are able to avoid the machine translation step (TRANSLATE-TEST) from the target language into English. The results also indicate that X-MAML boosts Multi-BERT performance on XNLI. It is worthwhile mentioning that Multi-BERT in the TRANSLATE-TRAIN setup outperforms few-shot X-MAML, however, we only use 2k development examples from the target languages, whereas in the aforementioned work, 433k translated sentences are used for fine-tuning.

\subsection{Zero-Shot Cross-Lingual QA}
We use a similar approach for cross-lingual QA on the MLQA dataset.
Zero-shot results on MLQA are shown in Table \ref{tab:mlqa_results}. %shows the results of zero-shot X-MAML for the MLQA dataset. 
We compare our results to those reported in two benchmark papers, \citet{hu2020xtreme} and \citet{Liang2020XGLUEAN}. We also report our own baselines for the task. The baselines are provided by training each base model on the SQuAD v1.1 train set (see Step~\ref{step1} in ~\secref{sec:mode_xmaml}) and evaluating on the test set of MLQA.
All target languages benefit from meta-learning with at least one of the auxiliary languages. %, highlighting the method's benefits. 
Using two auxiliary languages in X-MAML further improves results.
%Moreover, X-MAML with two auxiliary languages is more beneficial for the state-of-the-art models (\eg{XLM and XLM-R}) in cross-lingual QA.
Overall, zero-shot learning models with X-MAML outperform both internal and external baselines. %are improved compared to the results reported without meta-learning (\ie baselines). 
The improvement is $+1.04\%$, $+0.89\%$ and $+1.47\%$ in average F$_1$ score compared to XLM-15, XLM-R$_{base}$ and XLM-R$_{large}$, respectively. 
\addtolength{\tabcolsep}{+3pt} % hack to have thinner columns
\begin{table*}[t]
\centering
\fontsize{10}{10}\selectfont
\resizebox{.9\textwidth}{!}{
\begin{tabular}{c|c|l|ccccccc|c}
\toprule
\multicolumn{3}{c|}{Model} & en    & ar                   & de      & es                   & hi                   & vi                   & zh                   & avg                  \\
\toprule
\parbox[t]{6mm}{\multirow{6}{*}{\rotatebox{90}{XLM}}} & &
Our baseline & \textbf{69.80 }  &	48.95   & 52.64    &	\textbf{58.15 } &	46.67   & 48.46   & 42.64   & 52.47  \\

\cmidrule(r){2-11}
& \parbox[t]{4mm}{\multirow{4}{*}{\rotatebox{90}{\textbf{X-MAML}}}} &{{\it (One aux. lang.)}}  &  69.39      &	 48.45      &  53.04      &	57.68     &	 46.90      &  49.79     &  44.36     	& \multirow{2}{*}{52.80}  \\

& & $l\rightarrow X$ &  \textit{ar}  &	    \textit{hi}  & \textit{es}  & \textit{en} &	  \textit{zh}  &     \textit{zh} &     \textit{hi} 	&   \\
\cmidrule(r){3-11}
& &{{\it (Two aux. lang.)}}  & 68.88   &\textbf{49.76}   & \textbf{53.18}  &	58.00  &	\textbf{48.43}   & \textbf{50.86} & \textbf{45.44} 	&  \multirow{2}{*}{\textbf{53.51}} \\

& & $(l_1,l_2)\rightarrow X$ &  \textit{(es,ar)}  &  \textit{(vi,zh)}  &   \textit{(vi,zh)} &	  \textit{(en,zh)} &	  \textit{(vi,zh)}  &   \textit{(hi,zh)} & \ \textit{(es,hi)}	& \\
\midrule
\midrule
\parbox[t]{6mm}{\multirow{7}{*}{\rotatebox{90}{XLM-R$_{base}$}}} &&
\citet{Liang2020XGLUEAN} & {80.1  } & {56.4 } & {62.1 } & {67.9  } & {60.5 } & {67.1 } & {61.4 } & { 65.1 } \\ 
& & Our baseline & \textbf{80.38}  &	57.23  & 63.08 &	{67.91 } &	61.46   & 67.14  & 62.73  	& 65.70  \\
\cmidrule(r){2-11}
& \parbox[t]{3mm}{\multirow{4}{*}{\rotatebox{90}{\textbf{X-MAML}}}}& {\it (One aux. lang.)}  & {80.19 }  & {57.97 } & { 63.57 } & {67.46 } & {61.70 } & {67.97 }&  {64.01} & \multirow{2}{*}{66.12} \\ 

 && $l\rightarrow X$ & \textit{vi} &  \textit{hi}& \textit{ar}&  \textit{vi}&  \textit{vi}&  \textit{hi}&   \textit{hi}& \\ 
\cmidrule(r){3-11}
& &{\it (Two aux. lang.)} & 80.31  & \textbf{58.14}  & \textbf{64.07 }  &  \textbf{68.08 } & \textbf{62.67 }  & \textbf{68.82 }  & \textbf{64.06 } & \multirow{2}{*}{\textbf{66.59} } \\ 

&& $(l_1,l_2)\rightarrow X$ &  \textit{(ar,vi)} &   \textit{(hi,vi)}&   \textit{(ar,hi)}&    \textit{(ar,hi)}&   \textit{(es,ar)}&   \textit{(ar,hi) }&   \textit{(ar,hi)}&   \\ 
\midrule
\midrule
%\rowcolor{gray!20}\multicolumn{9}{l}{\it  Zero-shot cross-lingual transfer with XLM-R$_{large}$} \\
\parbox[t]{6mm}{\multirow{7}{*}{\rotatebox{90}{XLM-R$_{large}$}}}&&
\citet{hu2020xtreme}& {83.5 } & {66.6 } & {70.1 } & {74.1 } & {70.6 } & {74}   & {62.1} & {71.6 } \\

&& Our baseline & {83.95 }  &	66.09  & 70.62 &	{74.59 }  &	70.64  & 74.13  & 69.80	& 72.83 \\
\cmidrule(r){2-11}
&\parbox[t]{3mm}{\multirow{4}{*}{\rotatebox{90}{\textbf{X-MAML}}}}& {\it (One aux. lang.)} & 84.31   & {66.61}    & {70.84 }   & 74.32   & \textbf{70.94 }  & \textbf{74.84}   & \textbf{70.74} & \multirow{2}{*}{73.23} \\
 && $l\rightarrow X$ &  \textit{ar} &    \textit{hi} &   \textit{ar} &   \textit{hi} &  \textit{vi}&   \textit{ar} &  \textit{hi}&    \\
\cmidrule(r){3-11}
&& {\it (Two aux. lang.)} & \textbf{84.60}  & \textbf{66.95}  & \textbf{71.00} & \textbf{74.62}  & 70.93  & 74.73   & 70.29    & \multirow{2}{*}{\textbf{74.30}} \\ 

& &$(l_1,l_2)\rightarrow X$  &   \textit{(hi,vi)}&  \textit{(hi,vi)}&   \textit{(ar,vi)}&   \textit{(en,vi)}&   \textit{(ar,vi)}&   \textit{(es,hi)}&  \textit{(en,vi)} &  \\ 
\bottomrule
\end{tabular}}
\caption{F$_1$ scores (average over 10 runs) on the MLQA test set using zero-shot X-MAML. Columns indicate the target languages. The avg column indicates row-wise  average F$_1$ score. We also report the most beneficial auxiliary language(s) for X-MAML in improving the test F$_1$ of each target language.}
\label{tab:mlqa_results}
\end{table*}

We also evaluate on the less widely used cross-lingual QA dataset X-WikiRE \citep{abdou-etal-2019-x} for which we observe similar result trends, and $0.55\%$ improvement in terms of average F$_1$ score on zero-shot QA.
More details can be found in the Appendix (Section \ref{app:x-wikire}).

\section{Related Work}
The main motivation for this work is the low availability of labelled training datasets for most of the world's languages. To alleviate this issue, a number of methods, including the so-called few-shot learning approaches have been proposed.
Few-shot learning methods have initially been introduced within the area of image classification \cite{vinyals:16,ravi:17,finn:17}, but have recently also been applied to NLP tasks such as relation extraction \cite{han:18}, text classification \cite{yu:18,rethmeier2020longtail} and machine translation \cite{gu-etal-2018-meta}. Specifically, in NLP, these few-shot learning approaches include: (i) the transformation of the problem into a different task (\eg relation extraction is transformed to question answering~\cite{levy-etal-2017-zero,abdou-etal-2019-x}); or (ii) meta-learning \cite{andrychowicz:16,finn:17}.
\paragraph{Meta-Learning:}
Meta-learning or learning-to-learn has recently received a lot of attention from the NLP community.
First-order MAML has been applied to the task of machine translation \cite{gu-etal-2018-meta}, where they propose to use meta-learning for improving the machine translation performance for low-resource languages by learning to adapt to target languages based on multilingual high-resource languages. However, in the proposed framework, they include 18 high-resource languages as auxiliary languages and five diverse low-resource languages as target languages. In our work, we assume access to only English as a high-resource language.  
For the task of dialogue generation, \newcite{qian-yu-2019-domain} address domain adaptation using meta-learning.
\newcite{dou-etal-2019-investigating} explore MAML variants thereof for low-resource NLU tasks in the GLUE dataset \cite{DBLP:journals/corr/abs-1804-07461}. They consider different high-resource NLU tasks such as MultiNLI \cite{williams-etal-2018-broad} and QNLI \cite{DBLP:journals/corr/RajpurkarZLL16} as auxiliary tasks to learn meta-parameters using MAML. They then fine-tune the low-resource tasks using the adapted parameters from the meta-learning phase.
All the above-mentioned works on meta-learning in NLP assume that there are multiple high-resource tasks or languages, which are then adapted to new target tasks or languages with a handful of training samples. However, in a cross-lingual NLI and QA setting, the available high-resource language is usually only English. Our work thus fills an important gap in the literature, as we only require a single source language.
\paragraph{Cross-Lingual NLU:}
Cross-lingual learning has a fairly short history in NLP, and has mainly been restricted to traditional NLP tasks, such as PoS tagging, morphological inflection and parsing. In contrast to these tasks, which have seen much cross-lingual attention \cite{plank2016multilingual,bjerva:2017,kementchedjhieva-etal-2018-copenhagen,delhoneux:2018}, there has been relatively little work on cross-lingual NLU, partly due to lack of benchmark datasets. Existing work has mainly been focused on NLI \cite{agic2017baselines,conneau2018xnli,zhao2020inducing}, and to a lesser degree on RE \cite{faruqui2015multilingual,verga2015multilingual} and QA \cite{abdou-etal-2019-x,lewis2019mlqa}.
Previous research generally reports that cross-lingual learning is challenging and that it is hard to beat a machine translation baseline (\eg \newcite{conneau2018xnli}). Such a baseline is for instance suggested by \newcite{faruqui2015multilingual}, where the text in the target language is automatically translated to English.
We achieve competitive performance compared to a machine translation baseline (for XNLI), and propose a method that requires no training instances for the target task in the target language. Furthermore, our method is model agnostic, and can be used to extend any pre-existing model.
\section{Discussion and Analysis} 
\paragraph{Cross-Lingual Transfer:}
Somewhat surprisingly, we find that cross-lingual transfer with meta-learning yields improved results even when languages strongly differ from one another.
For instance, for zero-shot meta-learning on XNLI, we observe gains for almost all auxiliary languages, with the exception of Swahili (sw).
This indicates that the meta-parameters learned with X-MAML are sufficiently language agnostic, as we otherwise would not expect to see any benefits in transferring from, \eg Russian (ru) to Hindi (hi) (one of the strongest results in~\figref{fig:heat-map-mBERT-zero}).
This is dependent on having access to a pre-trained multilingual model such as BERT, however, using monolingual BERT (En-BERT) yields overwhelmingly positive gains in some target/auxiliary settings  (see additional results in~\figref{fig:heatmap-MAML-EBERT} in the Appendix).
For few-shot learning, our findings are similar, as almost all combinations of auxiliary and target languages lead to improvements when using Multi-BERT (\figref{fig:ffew-shot-heatmap-MAML-MBERT} in the Appendix).
However, when we only have access to a handful of training instances as in few-shot learning, even the English BERT model mostly leads to improvements in this setting (see additional results in~\figref{fig:few-shot-heatmap-MAML-EBERT} in the Appendix).

\paragraph{Typological Correlations:}
In order to better explain our results for cross-lingual zero-shot and few-shot learning, we investigate typological features, and their overlap between target and auxiliary languages.
We evaluate on the World Atlas of Language Structure (WALS, \newcite{wals}), which is the largest openly available typological database.
It comprises approximately 200 linguistic features with annotations for more than 2500 languages, which have been made by expert typologists through study of grammars and field work.
We draw inspiration from previous work~\cite{bjerva-augenstein-2018-phonology,bjerva_augenstein:18:iwclul,bjerva-etal-2019-probabilistic,bjerva-etal-2019-uncovering,bjerva-etal-2019-language,zhao2020inducing} which attempts to predict typological features based on language representations learned under various NLP tasks.
Similarly, we experiment with two conditions:
(i) we attempt to predict typological features based on the mutual gain/loss in performance using X-MAML; 
(ii) we investigate whether sharing between two typologically similar languages is beneficial for performance using X-MAML.
We train a simple logistic regression classifier per condition above, for each WALS feature.
In the first condition (i), the task is to predict the exact WALS feature value of a language, given the change in accuracy in combination with other languages.
In the second condition (ii), the task is to predict whether a main and auxiliary language have the same WALS feature value, given the change in accuracy when the two languages are used in X-MAML.
We compare with two simple baselines, one based on always predicting the most frequent feature value in the training set, and one based on predicting feature values with respect to the distribution of feature values in the training set.
We then investigate whether any features could be consistently predicted above baseline levels, given different test-training splits.
We apply a simple paired t-test to compare our models predictions to the baselines.
As we are running a large number of tests (one per WALS feature), we apply Bonferroni correction, changing our cut-off $p$-value from $p=0.05$ to $p=0.00025$.

We first investigate few-shot X-MAML, when using Multi-BERT, as reported in~\tabref{tab:xnli-mBERT-fewshot} (Appendix).
We find that languages sharing the feature value for WALS feature \textit{67A The Future Tense} are beneficial to each other.
This feature encodes whether or not a language has an inflectional marking of future tense, and can be considered to be a morphosyntactic feature.
We next look at zero-shot X-MAML with Multi-BERT, as reported in~\tabref{tab:xnli-mBERT-zeroshot} (Appendix).
For this case, we find that languages sharing a feature value for the WALS feature \textit{25A Locus of Marking: Whole-language Typology} typically help each other. This feature describes whether the morphosyntactic marking in a language is on the syntactic heads or dependents of a phrase.  For example en,
de, ru, and zh are `dependent-marking' in this feature. And if we look at the results in~\figref{fig:heat-map-mBERT-zero}, they have the largest mutual gains from each other during the zero-shot X-MAML.
In both cases, we thus find that languages with similar morphosyntactic properties can be beneficial to one another when using X-MAML. 

\section{Conclusion}
In this work, we show that meta-learning can be used to effectively leverage training data from an auxiliary language for zero-shot and few-shot cross-lingual transfer. 
We evaluated this on two challenging NLU tasks (NLI and QA), and on a total of 15 languages.
We are able to improve the performance of state-of-the-art baseline models for (i) zero-shot XNLI, and (ii) zero-shot QA on the MLQA dataset.
Furthermore, we show in a typological analysis that languages which share certain morphosyntactic features tend to benefit from this type of transfer.
Future studies will extend this work to other cross-lingual NLP tasks and more languages.
%with a more representative typological distribution.
%\setcounter{secnumdepth}{0}
\section*{Acknowledgements}
This research has received funding from the Swedish Research Council under grant agreement No  2019-04129, as well as the Research Foundation - Flanders (FWO).
This work was also funded by UiO: Energy to support international mobility. We are grateful to the Nordic Language Processing Laboratory (NLPL) for providing access to its supercluster infrastructure.
%use of the Saga or Puhti superclusters through the Nordic Language Processing Laboratory (NLPL)
%UiO: Energy grants for Copenhagen research visit

\bibliography{emnlp2020}
\bibliographystyle{acl_natbib}

\clearpage
\appendix

\section{Appendices}
\label{sec:appendix}
%\subsection{Experimental Setup}\label{setup}
\subsection{X-MAML using X-WikiRE dataset} \label{app:x-wikire}
\paragraph{X-WikiRE:}
\newcite{levy-etal-2017-zero} frame the Relation Extraction (RE) task as a QA problem using pre-defined natural language question templates. For example, a relation type such as \texttt{author} is transformed to at least one language question template (\eg \texttt{who is the author of x?}, where \texttt{x} is an entity). 
Building on the work of~\newcite{levy-etal-2017-zero}, a new dataset (X-WikiRE) is introduced for multilingual QA-based relation extraction in five languages (\ie English, French, Spanish, Italian and German) by \newcite{abdou-etal-2019-x}.
Each instance in the dataset includes a question, a context, and an answer. The question is a transformation of a target relation and the context may contain the answer. If the answer is not present, it is marked as \texttt{NIL}. In this task, we evaluate the performance of our method on the \texttt{UnENT} setting of the X-WikiRE dataset, where the goal is to generalise to unseen entities. For the evaluation, we use F$_1$ scores (for questions with valid answers) similar to~\newcite{kundu-ng-2018-nil}. \paragraph{QA experiments:}
We use the Nil-Aware Answer Extraction Framework (NAMANDA, \newcite{kundu-ng-2018-nil})\footnote{\url{https://github.com/nusnlp/namanda}} as the base model \texttt{M} in X-MAML for our QA experiments. NAMANDA encodes the question and context sequences to compute a similarity matrix. It creates evidence vectors through joint encoding of question and context and applies multi-factor self-attentive encoding. Finally, the evidence vectors are decomposed to output either the answer to the question or \texttt{NIL}.
We set the parameters to the default values (as in the original work) for the training and evaluation phases.
The NAMANDA model \texttt{M} is pre-trained on the full English training set (1M instances - see Step~\ref{step1}~in our training algorithm). The model \texttt{M} is further used by our meta-learning step to adapt the pre-trained QA model. We then evaluate how well the English model has been adapted by each of the auxiliary language through X-MAML via performing either few- or zero-shot learning.
In few-shot X-MAML, the meta-learned \texttt{M} is fine-tuned on the development set (1k instances) of other languages (\ie fr, es, it and de).
For both few- and zero-shot learning, we evaluate on the 10k test set of each of the target languages. Following the work of \newcite{abdou-etal-2019-x}, the Multi-BERT model is used to jointly encode text for different languages in the QA model.
\paragraph{Zero- and Few-Shot Cross-Lingual QA:}
We use a similar approach for cross-lingual QA on the X-WikiRE dataset.
Table \ref{tab:zero-few-X-WikiRE} shows the results of both zero- and few-shot X-MAML for the \texttt{UnENT} part (\ie generalise to unseen entities) of the X-WikiRE dataset. We compare our results for the \texttt{UnENT} scenario on the X-WikiRE dataset to those reported in the original paper. 
All of the target languages benefit from at least one of the auxiliary languages by adapting the model using X-MAML, highlighting the benefits of this method.
We were not able to directly reproduce the result for the zero-shot scenario of the original paper, thus we also report our own baseline for the task. We find that: (i) our zero-shot results with X-MAML improve on those without meta-learning (\ie baselines); and (ii) we outperform \newcite{abdou-etal-2019-x} for the \texttt{UnENT} scenario of zero-shot cross-lingual QA. 
Furthermore, for the few-shot scenario, adapting the QA model using few-shot X-MAML with only 1k development data outperforms their cross-lingual transfer model where \newcite{abdou-etal-2019-x} use 10k in the fine-tuning phase. 
\addtolength{\tabcolsep}{+3pt} % hack to have wider columns again
\begin{table*}[t]
\begin{center}
\resizebox{.9\linewidth}{!}{
\begin{tabular}{lc|rrrr|r|rrrr}
\toprule
& & \multicolumn{4}{c|}{Auxiliary language} & Baseline & \multicolumn{4}{c}{\newcite{abdou-etal-2019-x}} \\
\cmidrule(ll){8-11}
& {} &     es &     fr &     it &     de & &\multicolumn{2}{c}{BERT} & \multicolumn{2}{c}{fastText}   \\
\midrule
\parbox[t]{2mm}{\multirow{4}{*}{\rotatebox[origin=c]{90}{zero-shot}}} & es &   - &  49.01 &  \textbf{50.11} & \textbf{ 50.59} &      49.85 & \multicolumn{2}{c}{5.49} & \multicolumn{2}{c}{16.17} \\
&fr &  \textbf{52.20} &   - &  \textbf{52.13} &  \textbf{51.96} &      51.72 & \multicolumn{2}{c}{17.42} & \multicolumn{2}{c}{15.28} \\
&it &  50.53 &  \textbf{50.65} &   - &  50.58 &      50.58  & \multicolumn{2}{c}{10.70}& \multicolumn{2}{c}{4.44}\\
&de &  \textbf{49.92} &  48.78 &  48.63 &   - &      48.98 & \multicolumn{2}{c}{2.87} & \multicolumn{2}{c}{14.09}\\
\midrule
&  &    \multicolumn{5}{c|}{1k}  & 1k &10k &1k & 10k \\
\cmidrule(ll){3-7}
\cmidrule(ll){8-9}
\cmidrule(ll){10-11}
\parbox[t]{2mm}{\multirow{4}{*}{\rotatebox[origin=c]{90}{few-shot}}}&es &   - &  78.09 &  \textbf{78.33} &  77.89 &      78.26 & 42.97&71.66 & 65.78 & 77.99 \\
&fr &  \textbf{80.68} &   - &  \textbf{80.81} &  \textbf{80.74} &80.67 & 42.69 & 72.43 & 65.67 &74.15 \\
&it &  \textbf{82.04} &  81.76 &   - &  81.77 &      81.78 & 56.25 & 80.06 & 64.02 &\textbf{83.45}\\
&de &  78.29 &  78.48 &  \textbf{78.66} &   - &      78.63 & 56.01 &70.43 & 62.47 &72.17 \\
\bottomrule
\end{tabular}
}
\caption{F$_1$ scores (average over 10 runs) for the test set of the \texttt{UnENT} part of the X-WikiRE dataset using zero- and few-shot X-MAML. Baseline for (i) zero-shot learning: we evaluate the pre-trained NAMANDA model on the test set of the target language indicated in each row; and for (ii) few-shot learning:  we fine-tune the model on the development set and evaluate on the test set of the target language. We report results with few-shot X-MAML with only 1k instances from the development set.}
\label{tab:zero-few-X-WikiRE}
\end{center}
\end{table*}

\begin{figure*}[!ht]
  \centering
    \includegraphics[width=1\textwidth]{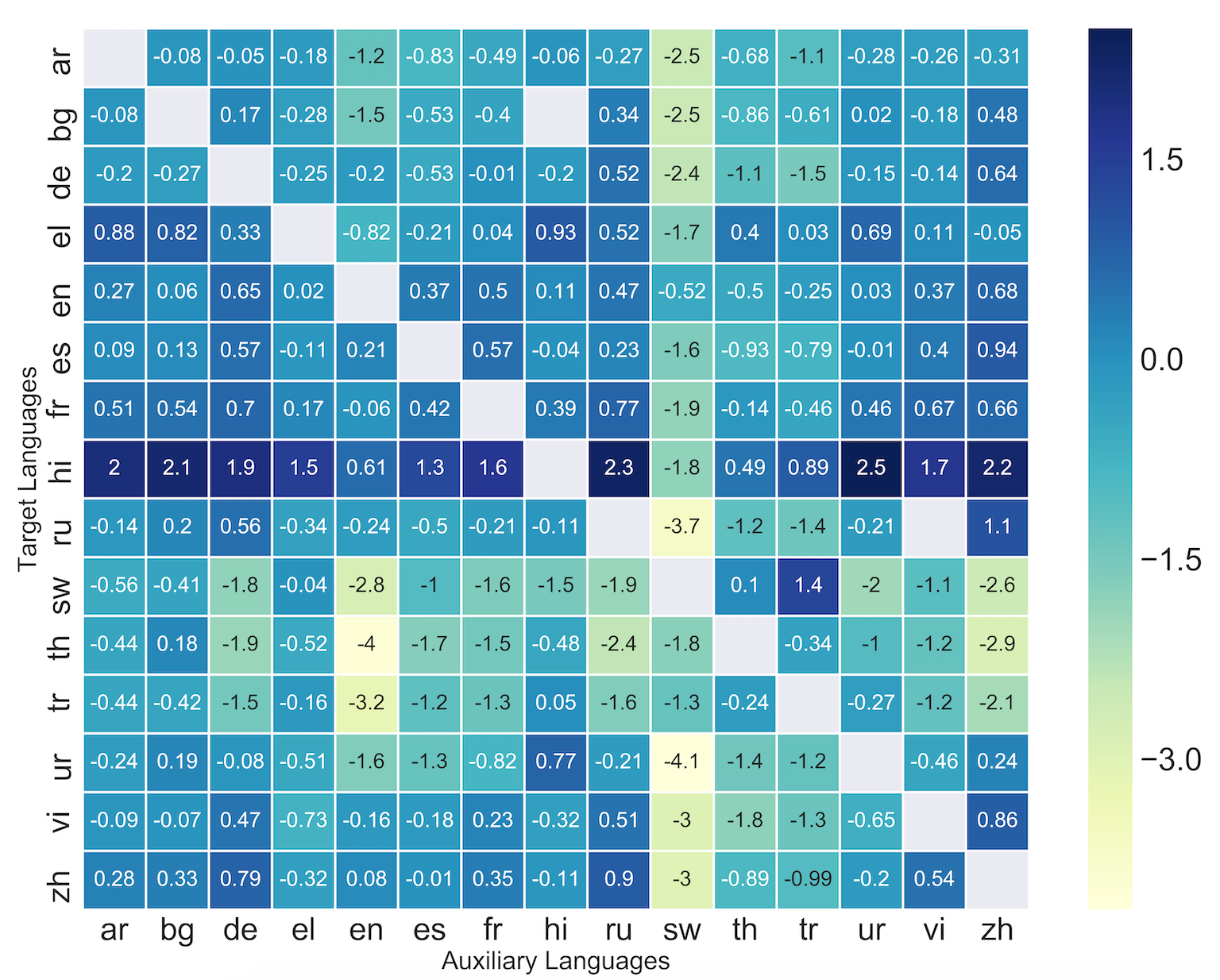}
    \caption{Differences in performance in terms of accuracy scores on the test set for the zero-shot case using training (without meta-learning) on XNLI with the Multi-BERT model. Rows correspond to target and columns to auxiliary languages. Numbers on the off-diagonal indicate performance differences between training on the auxiliary languages (without meta-learning) and the baseline model in the same row. The coloring scheme indicates the differences in performance (\eg blue for large improvement).
    }
    \label{fig:heatmap-NORMAL-MBERT}
\end{figure*}
\begin{figure*}[ht]
\centering
    \includegraphics[width=1\textwidth]{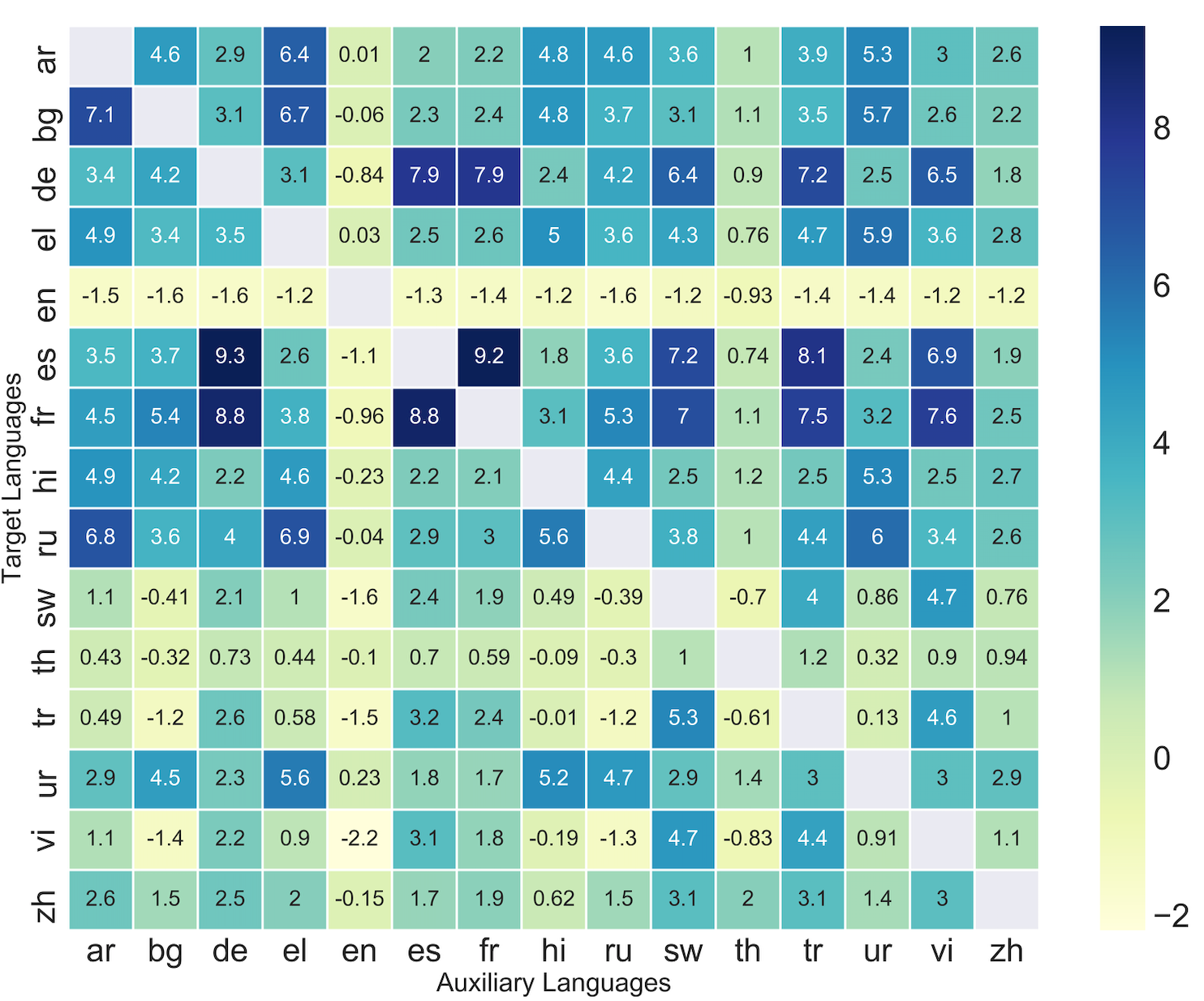}
    \caption{Differences in performance in terms of accuracy scores on the test set for zero-shot X-MAML on XNLI using the En-BERT (English) model. Rows correspond to target and columns to auxiliary languages used in X-MAML. Numbers on the off-diagonal indicate performance differences  between X-MAML and the baseline model in the same row. The coloring scheme indicates the differences in performance (\eg blue for large improvement).}
    \label{fig:heatmap-MAML-EBERT}
\end{figure*}
\begin{figure*}[ht]
  \centering
    \includegraphics[width=1\textwidth]{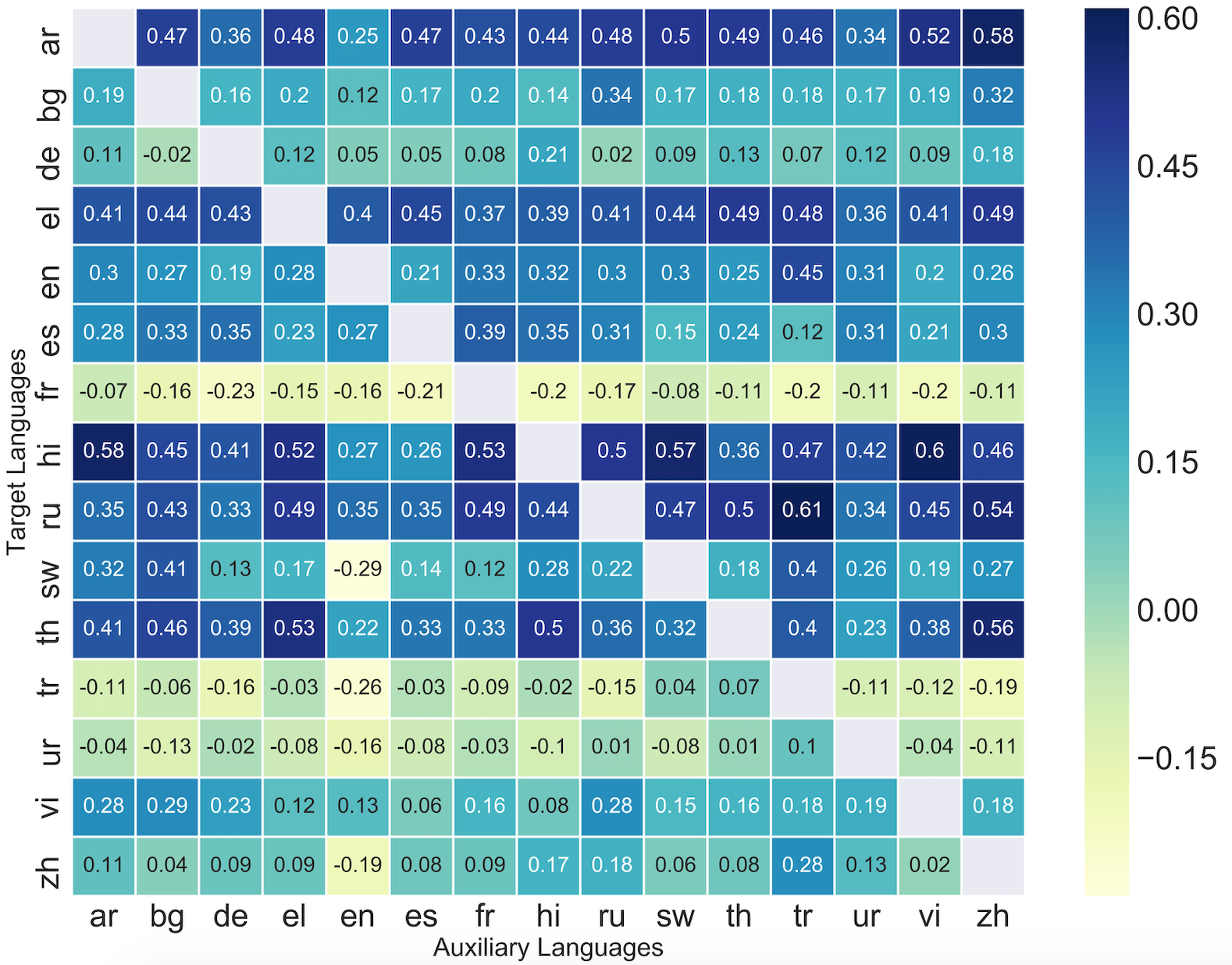}
    \caption{Differences in performance in terms of accuracy scores on the test set for few-shot X-MAML on XNLI using the Multi-BERT model. Rows correspond to target and columns to auxiliary languages used in X-MAML. Numbers on the off-diagonal indicate performance differences  between X-MAML and the baseline model in the same row. The coloring scheme indicates the differences in performance (\eg blue for large improvement).}
    \label{fig:ffew-shot-heatmap-MAML-MBERT}
  \end{figure*}
\begin{figure*}[ht]
    \includegraphics[width=1\textwidth]{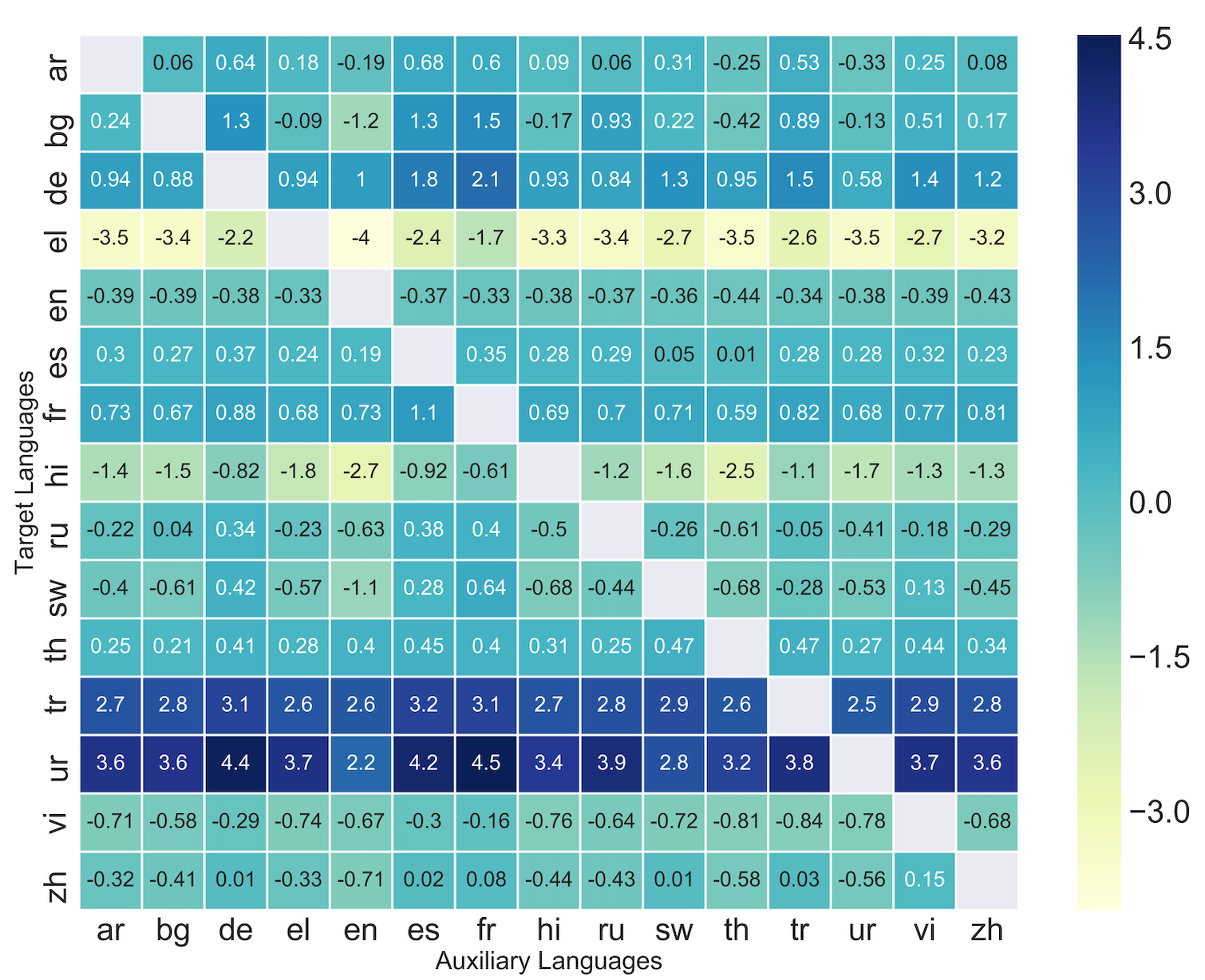}
    \caption{Differences in performance in terms of accuracy scores on the test set for few-shot X-MAML on XNLI using the En-BERT (English) model. Rows correspond to target and columns to auxiliary languages used in X-MAML. Numbers on the off-diagonal indicate performance differences  between X-MAML and the baseline model in the same row. The coloring scheme indicates the differences in performance (\eg blue for large improvement).}
    \label{fig:few-shot-heatmap-MAML-EBERT}
\end{figure*}

\addtolength{\tabcolsep}{-3pt} % hack to have thinner columns

%%%%% table for zero-shot xMAML mBERT XNLI
%\addtolength{\tabcolsep}{-3pt} % hack to have thinner columns
\begin{table*}[t]
\begin{center}
\resizebox{1\linewidth}{!}{
\begin{tabular}{lccccccccccccccc|r}
\toprule
 &\multicolumn{15}{c|}{\textbf{Auxiliary language}} & \textbf{baseline}\\
{} &     \textbf{ar} &     \textbf{bg} &     \textbf{de} &     \textbf{el} &     \textbf{en} &     \textbf{es} &     \textbf{fr} &     \textbf{hi} &     \textbf{ru} &     \textbf{sw} &     \textbf{th} &     \textbf{tr} &     \textbf{ur} &     \textbf{vi} &     \textbf{zh}  & \\
\midrule
\textbf{ar} &   - &  65.76 &  65.48 &  66.05 &  64.41 &  65.27 &  65.24 &  65.86 &  65.31 &  63.66 &  65.25 &  65.58 &  65.56 &  65.84 &  65.32 & 64.63\\
\textbf{bg} &  68.36 &   - &  68.79 &  68.39 &  67.95 &  68.45 &  68.80 &  68.86 &  69.41 &  66.10 &  67.62 &  67.95 &  68.63 &  68.67 &  69.45 & 67.82 \\
\textbf{de} &  70.88 &  71.46 &   - &  71.26 &  71.09 &  71.12 &  71.11 &  71.59 &  71.83 &  68.65 &  70.29 &  70.37 &  71.42 &  71.15 &  71.83 & 69.74 \\
\textbf{el} &  67.53 &  67.58 &  67.25 &   - &  66.11 &  67.13 &  67.39 &  67.95 &  67.71 &  65.11 &  67.12 &  67.15 &  67.69 &  67.19 &  67.34 & 65.73\\
\textbf{en} &  81.68 &  81.79 &  82.02 &  81.77 &   - &  81.88 &  81.91 &  81.88 &  82.03 &  80.44 &  81.18 &  81.43 &  81.80 &  81.73 &  82.09 & 81.36\\
\textbf{es} &  74.48 &  74.51 &  74.63 &  74.58 &  74.41 &   - &  74.95 &  74.81 &  74.63 &  72.66 &  73.91 &  74.12 &  74.51 &  74.71 &  75.07 & 73.85\\
\textbf{fr} &  74.13 &  74.02 &  74.22 &  74.11 &  73.75 &  74.18 &   - &  74.17 &  74.34 &  71.87 &  73.04 &  73.41 &  74.15 &  74.21 &  74.42 & 73.45\\
    \textbf{hi} &  60.75 &  61.59 &  60.84 &  60.61 &  59.31 &  60.18 &  60.66 &   - &  61.75 &  57.10 &  59.39 &  60.47 &  62.20 &  60.76 &  61.56 & 58.56\\
\textbf{ru} &  68.78 &  69.47 &  69.47 &  68.93 &  68.64 &  68.89 &  69.25 &  69.44 &   - &  66.11 &  68.18 &  68.72 &  69.52 &  69.02 &  70.19 & 67.94\\
\textbf{sw} &  48.71 &  48.53 &  47.36 &  49.13 &  46.70 &  48.43 &  47.81 &  47.11 &  47.28 &   - &  49.20 &  49.76 &  46.61 &  48.43 &  46.50 & 47.58\\
\textbf{th} &  54.65 &  55.39 &  53.80 &  54.98 &  51.14 &  54.09 &  54.15 &  55.26 &  53.82 &  52.90 &   - &  55.24 &  53.79 &  54.99 &  52.85 & 52.46 \\
\textbf{tr} &  60.94 &  61.20 &  60.22 &  61.09 &  58.66 &  60.60 &  60.32 &  60.93 &  60.29 &  59.98 &  60.53 &   - &  60.82 &  60.68 &  59.47 & 59.04\\
\textbf{ur} &  60.30 &  60.87 &  60.34 &  60.20 &  58.82 &  59.81 &  60.12 &  61.51 &  61.02 &  56.37 &  59.38 &  60.02 &   - &  59.87 &  60.46 & 58.70 \\
\textbf{vi} &  71.27 &  71.56 &  71.32 &  71.14 &  70.35 &  71.22 &  71.42 &  71.57 &  71.73 &  68.11 &  69.87 &  70.53 &  71.43 &   - &  71.82 & 70.12\\
\textbf{zh} &  70.24 &  70.68 &  70.65 &  70.12 &  69.91 &  70.29 &  70.47 &  70.59 &  71.11 &  67.47 &  69.33 &  69.50 &  70.29 &  70.54 &   -  & 68.90\\
\bottomrule
\end{tabular}
}
\caption{The performance in terms of average test accuracy for the zero-shot setting over 10 runs of X-MAML on the XNLI dataset using Multi-BERT (multilingual BERT), as base model. Each column corresponds to the performance of the Multi-BERT system after meta-learning with a single auxiliary language, and evaluation on the target language of the XNLI test set. The auxiliary language is not included during the evaluation phase. Results of the Multi-BERT model without X-MAML (baseline) are also reported.
}
\label{tab:xnli-mBERT-zeroshot}
\end{center}
\end{table*}

%\begin{table*}[]
%\resizebox{1\linewidth}{!}{
%\begin{tabular}{llllllllllllllll}
%Target Lang.          &\textbf{ar}    & \textbf{bg}    & \textbf{de}    & \textbf{el}    & \textbf{en}    & \textbf{es}    & \textbf{fr}    & \textbf{hi}    & \textbf{ru}    & \textbf{sw}    & \textbf{th}    & \textbf{tr}    & \textbf{ur}    &\textbf{vi}    & \textbf{zh}    \\
%\midrule
%Auxiliary Pair Lang. & \emph{el-tr} & \emph{hi-ru} & \emph{bg-zh} & \emph{ur-ru} & \emph{hi-de} & \emph{fr-de} & \emph{hi-ar} & \emph{ur-ru} & \emph{de-bg} & \emph{el-tr} & \emph{bg-tr} & \emph{ur-sw} & \emph{hi-de} & \emph{de-bg} & \emph{ru-el} \\
%\midrule
%Gain in Test Accuracy               & 2.56  & 3.6   & 3.71  & 3.43  & 1.23  & 2.12  & 2.24  & 4.97  & 3.5   & 2.84  & 4.37  & 3.53  & 4.23  & 2.51  & 4.23 
%\end{tabular}
%}
%\caption{Gains in test accuracies for Zero-shot X-MAML on XNLI using Multi-BERT and two auxiliary languages}
%\label{tbl:xnli-mBERT-zero-shot-pairlang}
%\end{table*}

\begin{table*}[th]
\begin{center}
\resizebox{1\linewidth}{!}{
\begin{tabular}{lccccccccccccccc|r}
\toprule
 &\multicolumn{15}{c|}{\textbf{Auxiliary language}} & \textbf{baseline}\\
{} &     \textbf{ar} &     \textbf{bg} &     \textbf{de} &     \textbf{el} &     \textbf{en} &     \textbf{es} &     \textbf{fr} &     \textbf{hi} &     \textbf{ru} &     \textbf{sw} &     \textbf{th} &     \textbf{tr} &     \textbf{ur} &     \textbf{vi} &     \textbf{zh}  & \\
\midrule
\textbf{ar} &   - &  67.84 &  67.73 &  67.85 &  67.62 &  67.84 &  67.80 &  67.81 &  67.85 &  67.87 &  67.86 &  67.83 &  67.71 &  67.89 &  67.95 & 67.37 \\
\textbf{bg} &  71.79 &   - &  71.76 &  71.80 &  71.72 &  71.77 &  71.80 &  71.74 &  71.94 &  71.77 &  71.78 &  71.78 &  71.77 &  71.79 &  71.92 & 71.60\\
\textbf{de} &  73.36 &  73.23 &   - &  73.37 &  73.30 &  73.30 &  73.33 &  73.46 &  73.27 &  73.34 &  73.38 &  73.32 &  73.37 &  73.34 &  73.43 & 73.25 \\
\textbf{el} &  69.95 &  69.98 &  69.97 &   - &  69.94 &  69.99 &  69.91 &  69.93 &  69.95 &  69.98 &  70.03 &  70.02 &  69.90 &  69.95 &  70.03 & 69.54 \\
\textbf{en} &  82.24 &  82.21 &  82.13 &  82.22 &   - &  82.15 &  82.27 &  82.26 &  82.24 &  82.24 &  82.19 &  82.39 &  82.25 &  82.14 &  82.20& 81.94 \\
\textbf{es} &  76.07 &  76.12 &  76.14 &  76.02 &  76.06 &   - &  76.18 &  76.14 &  76.10 &  75.94 &  76.03 &  75.91 &  76.10 &  76.00 &  76.09 &75.79\\
\textbf{fr} &  75.32 &  75.23 &  75.16 &  75.24 &  75.23 &  75.18 &   - &  75.19 &  75.22 &  75.31 &  75.28 &  75.19 &  75.28 &  75.19 &  75.28 & 75.39\\
\textbf{hi} &  64.95 &  64.82 &  64.78 &  64.89 &  64.64 &  64.63 &  64.90 &   - &  64.87 &  64.94 &  64.73 &  64.84 &  64.79 &  64.97 &  64.83 & 64.37\\
\textbf{ru} &  71.19 &  71.27 &  71.17 &  71.33 &  71.19 &  71.19 &  71.33 &  71.28 &   - &  71.31 &  71.34 &  71.45 &  71.18 &  71.29 &  71.38& 70.84\\
\textbf{sw} &  58.14 &  58.23 &  57.95 &  57.99 &  57.53 &  57.97 &  57.94 &  58.10 &  58.04 &   - &  58.00 &  58.22 &  58.08 &  58.01 &  58.09 &57.82\\
\textbf{th} &  61.59 &  61.64 &  61.57 &  61.71 &  61.40 &  61.51 &  61.51 &  61.68 &  61.54 &  61.50 &   - &  61.58 &  61.41 &  61.56 &  61.74 & 61.18\\
\textbf{tr} &  64.74 &  64.79 &  64.69 &  64.82 &  64.59 &  64.82 &  64.76 &  64.83 &  64.70 &  64.89 &  64.92 &   - &  64.74 &  64.73 &  64.66 &64.85 \\
\textbf{ur} &  63.67 &  63.58 &  63.69 &  63.63 &  63.55 &  63.63 &  63.68 &  63.61 &  63.72 &  63.63 &  63.72 &  63.81 &   - &  63.67 &  63.60 &63.71\\
\textbf{vi} &  73.51 &  73.52 &  73.46 &  73.35 &  73.36 &  73.29 &  73.39 &  73.31 &  73.51 &  73.38 &  73.39 &  73.41 &  73.42 &   - &  73.41 &73.23 \\
\textbf{zh} &  74.04 &  73.97 &  74.02 &  74.02 &  73.74 &  74.01 &  74.02 &  74.10 &  74.11 &  73.99 &  74.01 &  74.21 &  74.06 &  73.95 &   - &73.93 \\
\bottomrule
\end{tabular}
}
\caption{The performance in terms of average test accuracy for the few-shot setting over 10 runs of X-MAML on the XNLI dataset using Multi-BERT (multilingual BERT), as base model. Each column corresponds to the performance of the Multi-BERT system after meta-learning with a single auxiliary language, and evaluation on the target language of the XNLI test set. The auxiliary language is not included during the evaluation phase. Results of the Multi-BERT model without X-MAML (baseline) are also reported.}
\label{tab:xnli-mBERT-fewshot}
\end{center}
\end{table*}

\end{document}